\documentclass[letterpaper, 10 pt, conference]{ieeeconf}  

\usepackage{graphicx} 
\usepackage{amsmath}
\DeclareMathOperator*{\argmax}{argmax}
\IEEEoverridecommandlockouts                              

\usepackage{xcolor}
\usepackage{algorithm}
\usepackage{algorithmicx}
\usepackage{algpseudocode}
\usepackage{hyperref}
\usepackage{amssymb}
\usepackage{algpseudocode}
\usepackage{multicol}
\usepackage{subcaption}
\usepackage{graphicx}
\usepackage{lipsum}
\usepackage{soul}
\usepackage{cite}
\usepackage{tabularx}
\usepackage{array}
\usepackage{multirow}
\usepackage{fancyhdr}

\newcolumntype{L}{>{\raggedright\arraybackslash}X}
\algrenewcommand\algorithmicindent{0.8em}%

\algtext*{EndIf}

\DeclareMathOperator*{\argmin}{argmin}

\title{\LARGE \bf
Generalizing Cooperative Eco-driving via Multi-residual Task Learning
}

\author{Vindula Jayawardana$^{1^{*}}$, Sirui Li$^{1}$, Cathy Wu$^{1}$,
Yashar Farid$^{2}$, and Kentaro Oguchi$^{2}$
\thanks{$^{1}$Massachusetts Institute of Technology, Cambridge, MA, USA
{\tt\small \{vindula, siruil, cathywu\}@mit.edu}} 
\thanks{$^{2}$ Toyota InfoTech Labs and Toyota Motor North America, Mountain View, CA, USA {\tt\small \{yashar.zeiynali.farid, kentaro.oguchi\}@toyota.com} }%
\thanks{$^{*}$ Work done during the author's internship at Toyota InfoTech Labs.}
}

\begin{document}
\maketitle
\thispagestyle{empty}

\begin{abstract}
Conventional control, such as model-based control, is commonly utilized in autonomous driving due to its efficiency and reliability. However, real-world autonomous driving contends with a multitude of diverse traffic scenarios that are challenging for these planning algorithms. Model-free Deep Reinforcement Learning (DRL) presents a promising avenue in this direction, but learning DRL control policies that generalize to multiple traffic scenarios is still a challenge. To address this, we introduce Multi-residual Task Learning (MRTL), a generic learning framework based on multi-task learning that, for a set of task scenarios, decomposes the control into nominal components that are effectively solved by conventional control methods and residual terms which are solved using learning. We employ MRTL for fleet-level emission reduction in mixed traffic using autonomous vehicles as a means of system control. By analyzing the performance of MRTL across nearly 600 signalized intersections and 1200 traffic scenarios, we demonstrate that it emerges as a promising approach to synergize the strengths of DRL and conventional methods in generalizable control.

\end{abstract}

\section{INTRODUCTION}

Autonomous vehicles (AVs) are surging in popularity due to rapid technological advancements. Lately, AVs also have been used as Lagrangian actuators for system-level traffic control. Lagrangian control describes microscopic level traffic control techniques based on mobile actuators (e.g., vehicles) rather than fixed-location actuators (e.g., traffic signals). Therefore, it involves planning a fleet of AVs to accomplish a given objective at the system level, which involves both AVs and human drivers. These objectives include mitigating traffic congestion~\cite{flow}, curbing emissions~\cite{eco-drive}, and promoting smoother traffic flows~\cite{9811912}.

\begin{figure}
  \includegraphics[width=0.5\textwidth]{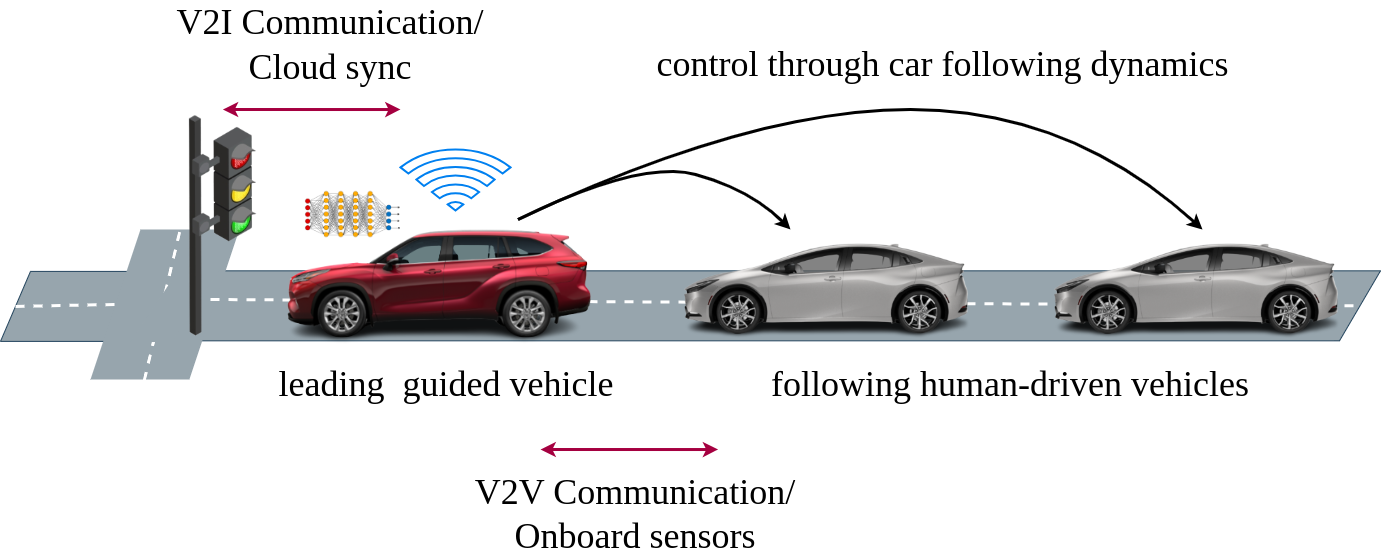} 
  \caption{In a signalized intersection, AVs lead platoons of human-driven vehicles. As Lagrangian actuators, they reduce fleet emissions by controlling their own acceleration and shepherding the human drivers through car following dynamics. \vspace*{-0.5cm}}
  \label{fig:schematics}
\end{figure}

In particular, recent work explores the use of AVs as Lagrangian actuators for fleet-wide emission reduction~\cite{huang2018eco}. As illustrated in Figure~\ref{fig:schematics}, the goal is to reduce emissions of the fleet by controlling and coordinating AVs and exerting control over human-driven vehicles. Methods from heuristics~\cite{katsaros2011performance}, to model-based methods~\cite{sajadi2019nonlinear}, to model-free~\cite{eco-drive} methods, have been used in tackling this challenge.

Although model-based control strategies like model-predictive control are frequently employed~\cite{huang2018eco}, they rely on the assumption of having a precise vehicle dynamics model. Yet, devising such a model, including the multitude of factors influencing driving dynamics, is challenging. In the absence of them, these methods often fall short in terms of adapting to various traffic scenarios. The deployment of simplified models tends to result in poor generalization and, in some cases, even increasing emissions levels. 

On the other hand, DRL operates without the need for a predefined dynamics model, hence model-free. It has the capacity to address control challenges that prove challenging for conventional methods as it specifies control objectives indirectly within a reward function rather than explicit control actions. Although limited in success, DRL has demonstrated the capacity to adapt to changes in the underlying environmental conditions~\cite{cobbe2019quantifying}, underscoring its potential to generalize across various traffic scenarios.

However, developing DRL algorithms for Lagrangian control in diverse traffic settings still remains a challenge. Real-world roads are complex, with complexities including vehicle interactions, varied road topologies, and external controls like traffic signals and stop signs. These complexities are often unpredictable, introducing uncertainty. Devising eco-driving planning algorithms capable of handling multiple scenarios is thus demanding, and many existing studies focus on a few scenarios~\cite{huang2018eco}, leading to less meaningful insights and potential overfitting in evaluations~\cite{jayawardana2022impact}.

In this study, we address this challenge of algorithmic generalization for DRL across various scenarios, such as different traffic scenarios in cooperative eco-driving. We introduce \textit{Multi-residual Task Learning (MRTL)}, a generic framework that combines DRL and conventional control methods. MRTL divides each scenario into parts solvable by conventional control and residuals solvable by DRL. The final control input for a scenario is thus the superposition of these two control inputs. 

We employ MRTL in eco-driving at signalized intersections, achieving better generalization in nearly 600 signalized intersections across 1200 traffic scenarios. While many existing works focus on a few eco-driving scenarios~\cite{huang2018eco}, to the best of our knowledge, we are the first to solve this problem on a large scale. 

Our key contributions are:

\begin{itemize}
    \item We present a generic learning framework called Multi-residual Task Learning to enable algorithmic generalization of DRL algorithms. 
    \item We employ the MRTL framework to devise generalizable control policies for cooperative eco-driving.   
    \item Analyzing nearly 1200 traffic scenarios across 600 signalized intersections, we demonstrate that our MRTL framework enables control generalization, outperforming baseline control methods by a large margin. 
\end{itemize}

\section{Related Work}

The use of residuals in learning has previously been explored both in supervised learning and reinforcement learning. He et al.~\cite{he2016deep} first propose residual neural networks where they reformulate the layers of a neural network as learning residual functions. In reinforcement learning, the use of residuals takes a slightly different form. The closest to our work is residual reinforcement learning (RRL), which was first introduced simultaneously by Silver et al.~\cite{silver2018residual} and Johannink et al.~\cite{sergey_rpl} for robot control.

Silver et al.~\cite{silver2018residual} primarily look at how RRL can improve the performance of robotics manipulation tasks with various nominal policies as backbones. They demonstrate that RRL performs well when the environment is partially observable and when there is sensor noise, model misspecifications, and controller miscalibrations. Johannink et al.~\cite{sergey_rpl} further show that RRL can be used to deal with the sim-to-real transfer. In particular, it can be used to modify the learned controller such that it can react to the modeling inaccuracies observed in the simulations to better adapt to the real world.

While these works lay a foundation, the focus is single-agent, single-task robotic manipulations characterized by relatively simple control. In contrast, we look at multi-agent, multi-task scenarios, necessitating not only optimizing performance on individual tasks but also extending to robust and generalization control synthesis.

In autonomous driving, Zhang et al.~\cite {zhang2022racing} use RRL and reduce the lap time in autonomous racing. They further show the transferability of learned policies by transferring a policy from one track to another and to new tracks. However, this work primarily looks at single-agent racing, and multi-agent racing poses different challenges. Furthermore, while some transferability results have been shown, it is still limited to a few select racing tracks instead of a diverse set of scenarios. Moreover, autonomous racing and Lagrangian control have contrasting dynamics: racing involves competition, while Lagrangian control involves cooperation.

RRL is further utilized in synthesizing generalizable reinforcement learning controllers for robotics manipulations. Hao et al.~\cite{hao2022metarpl} introduce a meta-residual policy learning method that performs in unseen peg-in-hole assembly tasks. It improves adaptation and sample efficiency but is limited to specific robotics skills and lacks task diversity. Further, it operates in low-dimensional state spaces, leaving its suitability for high-dimensional spaces uncertain. It's worth noting that our approach differs fundamentally, focusing on multi-task learning compared to their meta-learning approach. Nevertheless, this work underscores the potential of RRL in enhancing control policy generalization.

On the other hand, combining model-based methods with learning (model-free) has been a topic of interest for some time~\cite{hewing2020learning}. On the one hand, these methods often involve learning a dynamics model and then using the model for trajectory optimization or model predictive control~\cite{mordatch2016combining, kollar2018mpc} to simulate experience~\cite{deisenroth2013gaussian} or to compute gradients for model-free updates~\cite{heess2015learning}. In another line of work, learning is used for capturing the objectives~\cite{menner2019inverse} or constraints~\cite{chou2020learning}. Recently, there has also been a line of work that looks at learning a corrective term to analytical physical models~\cite{ajay2018augmenting} with the purpose of performing better predictive control. 

In summary, while RRL has proven effective in single-agent, single-task robotic manipulations, none of the existing studies have showcased its application to multi-agent cooperative control with the capacity for generalization across a range of scenarios, let alone in Lagrangian control. Alternatively, combining model-based and model-free methods has exhibited mixed results, and none of them adequately tackle the challenges of algorithmic generalization. In this study, we aim to bridge this gap in the field.

\section{Preliminaries}

\subsection{Reinforcement Learning}
\label{rl}

In reinforcement learning, an agent learns a control policy by interacting with its environment, typically modeled as a Markov Decision Process (MDP) denoted as $M = \left\langle\mathcal{S}, \mathcal{A}, p, r, \rho, \gamma \right\rangle$. Here, $\mathcal{S}$ represents the set of states, $\mathcal{A}$ denotes the possible actions, $p(s_{t+1}|s_t,a_t)$ denotes the transition probability from the current state $s_t$ to the next state $s_{t+1}$ upon taking action $a_t$,  the reward received for action $a_t$ at state $s_t$ is $r(s_t, a_t) \in \mathbb{R}$, a distribution over the initial states is $\rho$, and $\gamma \in [0,1]$ is a discounting factor that balances immediate and future rewards.

Given the MDP, we seek to find an optimal policy $\pi^*: \mathcal{S} \rightarrow \mathcal{A}$ over the horizon $H$ that maximizes the expected cumulative discounted reward over the MDP. 
\vspace*{-0.02cm}
\begin{equation}
\pi^*(s) = \argmax_{\pi} \mathop{\mathbb{E}} \left[\sum_{t=0}^{H}{\gamma^t r (s_t,a_t)| s_0, \pi}\right]
\end{equation}

\subsection{Multi-task Reinforcement Learning}
\label{multi-task-learning}

In multi-task reinforcement learning, we extend the single-MDP (single task) reinforcement learning in Section~\ref{rl} to multiple MDPs (multiple tasks). Accordingly, our objective in finding optimal policy thus becomes, 
\vspace*{-0.1cm}
\begin{equation}
\label{eq_mtl}
\pi^*(s) = \argmax_{\pi} \mathop{\mathbb{E}} \left[\sum_{\tau \in \mathcal{T}}\sum_{t=0}^{H}{\gamma^t r_{\tau}(s_t,a_t)| s_0, \pi}\right]
\end{equation}

where $\mathcal{T}$ is the set of MDPs (tasks). Also, note that what we seek in multi-task reinforcement learning is a unified policy that is performant over all MDPs (tasks). 

\section{Method}

In this section, we formalize the concept of algorithmic generalization in DRL and detail our generic Multi-Residual Task Learning framework. 

\subsection{Problem Formulation}

In this work, we study the algorithmic generalization of DRL algorithms across a family of MDPs (scenarios) that originate from a single task, such as eco-driving. To formalize this exploration, our primary focus revolves around solving Contextual Markov Decision Processes (cMDPs)~\cite{benjamins2022contextualize}.

A cMDP expands upon the MDP discussed in Section~\ref{rl} by incorporating a 'context'. Context serves as a means to parameterize the environmental variations encountered, such as changes in lane lengths at different intersections, among other factors in eco-driving. Mathematically, we denote a cMDP as $\mathcal{M} = \left\langle\mathcal{S}, \mathcal{A}, \mathcal{C}, p_c, r_c, \rho_c, \gamma \right\rangle$. Compared to MDPs, a context space $\mathcal{C}$ is introduced, and the action space $\textit{A}$ and state space $\textit{S}$ remain unchanged. The transition dynamics $\textit{p}_c$, rewards $\textit{r}_c$, and initial state distribution $\rho_c$ are changed based on the context $c \in \mathcal{C}$. Essentially, a cMDP $\mathcal{M}$ defines a collection of MDPs, each differing based on the context, such that $\mathcal{M} = \{M_c\}_{c \sim \mathcal{C}}$.

Solving a given cMDP leads to solving the problem of algorithmic generalization within that task (i.e., finding a policy that performs well in the cMDP overall). The generalization objective where the goal is to find a unified policy $\pi^*(\cdot)$ that performs well on all $M_c \in \mathcal{M}$ is as follows.
\vspace*{-0.2cm}
\begin{equation}
\label{eq-mrtl}
\pi^*(s) = \argmax_{\pi} \mathop{\mathbb{E}} \left[\sum_{c \in \mathcal{C}}\sum_{t=0}^{H}{\gamma^t r_{c}(s_t,a_t)| s_0^c, \pi}\right]
\end{equation}

The multi-task learning framework introduced in Section~\ref{multi-task-learning} emerges as a natural approach to tackle cMDPs. Here, the contexts themselves define the different tasks, effectively aligning with the notion that a specific context $c \in \mathcal{C}$ in Equation~\ref{eq-mrtl} corresponds to a task $\tau \in \mathcal{T}$ in Equation~\ref{eq_mtl}.

\subsection{Cooperative Eco-driving cMDP}
In cooperative eco-driving at signalized intersections, a wide array of context factors come into play, including lane lengths, speed limits, lane count, vehicle inflows, and the timings of green and red traffic signals. These factors collectively shape the contexts within the eco-driving cMDP, which encompasses a spectrum of signalized intersections (MDPs). Then, we seek a unified AV control policy that adeptly curbs emissions of the fleet across these signalized intersections.

MDPs within a cMDP can manifest in both single-agent and multi-agent configurations. However, cooperative eco-driving adopts multi-agent control as coordination between AVs to reduce emissions is required. This characteristic amplifies the complexity of solving eco-driving cMDP, necessitating the implicit modeling of vehicle interactions and addressing the challenges posed by partial observability. 

The overall objective of the cooperative eco-driving at signalized intersections is to minimize the emissions of a fleet of vehicles (both AVs and human-driven vehicles) while having a minimal impact on travel time across all signalized intersections. Given an instantaneous emission model $E(\cdot)$ that measures vehicular emission, we seek an AV control policy such that, 
\vspace*{-0.05cm}
\begin{equation}
\label{eco-drive-objective}
\pi^* = \argmin_{\pi} \mathop{\mathbb{E}} \left[\sum_{c \in \mathcal{C}} \sum_{i=1}^{n_c} \int_{0}^{T_i} E\left(a_i(t), v_i(t)\right) dt + T_i \right]
\end{equation}

Here, $n_c$ represents the total number of indexes of both AV and human-driven vehicles in intersection defined by $c$, $T_i$ denotes the travel time of vehicle $i$, $v_i(t)$, and $a_i(t)$ denote the speed and acceleration of vehicle $i$ at time $t$ and $\mathcal{C}$ denote the contexts defining a set of signalized intersections.

\subsection{Multi-residual Task Learning}

\begin{figure}[t]
\centering
\includegraphics[width=\linewidth]{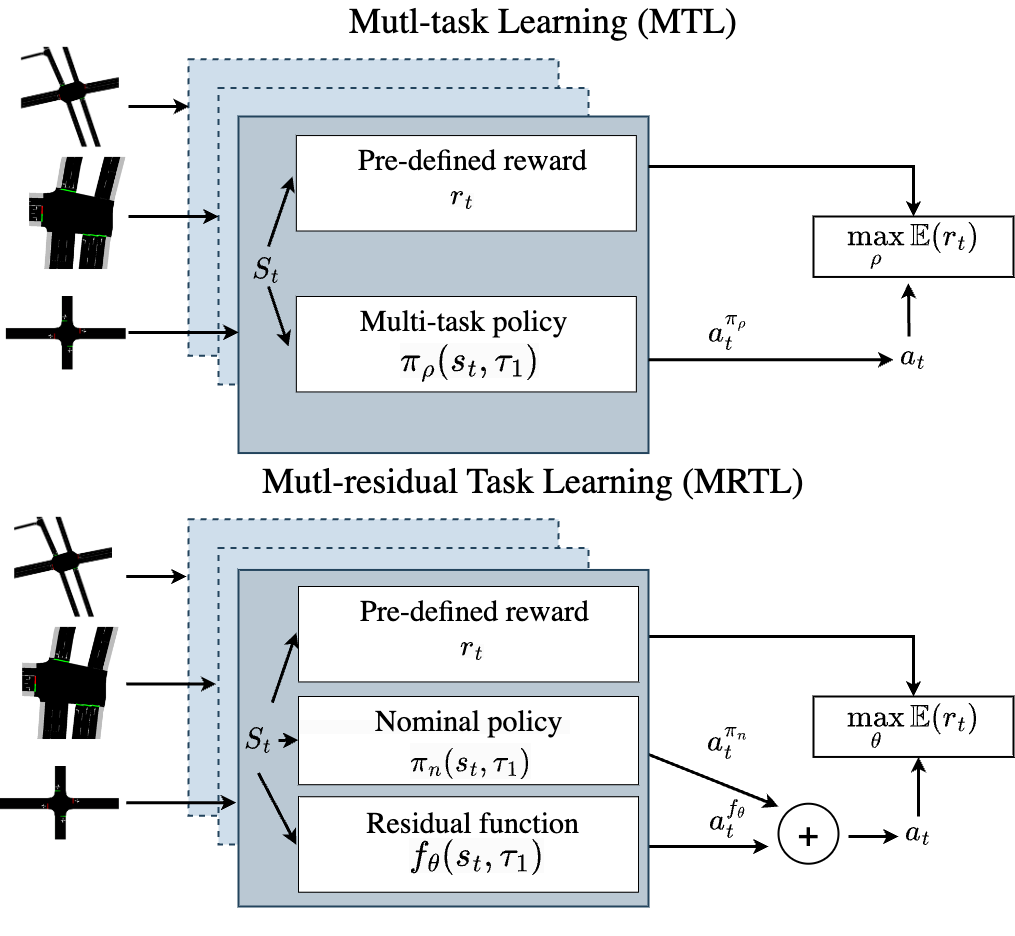}
\caption{Multi-task learning trains a unified policy directly with environments (intersections) sampled from a distribution of environments (top figure). Multi-residual task learning building on multi-task learning decomposes the cMDP into parts solved by a nominal policy and residual parts solved by DRL, as shown in the bottom figure. \vspace*{-0.4cm}}
\label{mrtl}
\centering
\end{figure}

While multi-task reinforcement learning can be used for solving cMDPs, it struggles when multiple MDPs are combined within one learning framework. Simultaneous training can lead to competition among MDPs for the learning agent's limited capacity, making it hard to balance MDP-specific and shared knowledge. Moreover, varying dynamics across MDPs challenge robust adaptation and generalization. Catastrophic interference risk, where new learning disrupts prior performance, hinders effectiveness further.

In addressing these issues, we propose a generic learning framework designed to enhance the algorithmic generalization of DRL algorithms, ultimately enabling solving cMDPs. We introduce Multi-Residual Task Learning, a unified learning approach that harnesses the synergy between multi-task learning and residual reinforcement learning~\cite{sergey_rpl, silver2018residual} (Figure~\ref{mrtl}). At its core, MRTL operates by augmenting any given nominal policy, which exhibits sub-optimal and varying performance in each MDP in a cMDP, by learning residuals on top of it. These residuals serve as corrective measures to address suboptimalities within the nominal policy.

Consider eco-driving at signalized intersections. An AV's overall emission-reduction reward, $r$, can be split into $r_a$ and $r_b$. $r_a$ rewards AV gliding at red signals, a known approach for emission reduction~\cite{huang2018eco}, and can be achieved through a straightforward model-based controller~\cite{katsaros2011performance}. Conversely, $r_b$ rewards adaptive gliding behaviors to environmental changes (e.g., other vehicles lane changing), a challenge for model-based controllers due to system complexity. Training a DRL policy for $r_b$ is feasible, allowing the nominal policy to fix $r_a$ while learning targets the residual $r_b$.

To put this formally, MRTL is concerned with augmenting a given nominal policy $\pi_n(s, c): \mathcal{S} \times \mathcal{C} \rightarrow \mathcal{A}$ by learning residuals on top of it. In particular, we aim to learn the MRTL policy $\pi(s, c): \mathcal{S}\times \mathcal{C} \rightarrow \mathcal{A}$ by learning a residual function $f_\theta(s, c): \mathcal{S} \times \mathcal{C} \rightarrow \mathcal{A}$ on top of a given nominal policy $\pi_n(s, c): \mathcal{S} \times \mathcal{C} \rightarrow \mathcal{A}$ such that, 
\vspace{-0.02cm}
\[
\pi(s, c) = \pi_n(s, c) + f_\theta(s, c)
\]

where $s \in \mathcal{S}$ and $c \in \mathcal{C}$. The gradient of the $\pi$ does not depend on the $\pi_n$. This enables flexibility with nominal policy choice. The effectiveness of $\pi_n$ can vary among different MDPs within a cMDP. Hence, the role of the residual function $f_\theta$ in each MDP depends on MDP characteristics and nominal policy performance in that MDP. In some MDPs, $\pi_n$ acts as a starting point for better exploration for the residual function. In others, it can be nearly optimal, requiring fewer improvements by the residual function.

\section{MRTL for Cooperative Eco-driving}

In this section, we discuss the application of the MRTL framework on eco-driving at signalized intersections. We procedurally generate a synthetic dataset with nearly 600 MDPs, which represent incoming approaches at signalized intersections. We simplify the eco-driving task to focus on these incoming approaches since traffic signals coordinate conflicting approaches~\cite{eco-drive}. 

Approaches are described by six features with diverse ranges: lane length (75-400 m), vehicle inflow (675-900 veh/hour), speed limit (10-15 m/s), lane count (1-3), and green and red signal phase times (25-30s). These features define the context space for the eco-driving cMDP, and each environment is a realization of these features.

\subsection{Nominal Policy}

As the nominal policy, we design a model-based heuristic controller inspired by the GLOSA algorithm for eco-driving~\cite{katsaros2011performance}. While our nominal policy doesn't perform real-time optimizations, its low computational demands and pre-deployment verification appeal to practical applications. We detail the nominal policy in Algorithm~\ref{nominal_policy}.

\begin{algorithm}[htb]
    \caption{Nominal policy $\pi_n$ for eco-driving}
    \label{nominal_policy}
    \begin{algorithmic}[1]
        \Procedure{Glide or Keep Speed}{ego-vehicle speed $v(t)$, ego-vehicle distance to intersection $d(t)$, traffic signal timing plan $T$ and green light duration $T_g$}
        \State Calculate time to intersection $T_I \gets \frac{d(t)}{v(t)}$
        \State Calculate time to green light $T_G$ from $T$
        \State Calculate time to end green light $T_E \gets T_G + T_g$
        \If {$T_G \leq T_I \leq T_E$}
            \State Target speed $v_{target} \gets v(t)$
        \ElsIf {$T_G \geq T_I$}
            \State Calculate target speed based on gliding principle\\
            \hspace*{15pt} $v_{target} \gets \frac{d(t)}{T_G}$
        \Else 
            \State Target speed $v_{target} \gets v_{IDM}$
        \EndIf
        \State \textbf{return} $v_{target}$
        \EndProcedure
    \end{algorithmic}
\end{algorithm}

The nominal policy operates on a simple set of criteria aimed at reducing idling 
and thereby reducing emissions~\cite{huang2018eco}. First, it checks if the ego-vehicle can pass the intersection at its current speed; if yes, it maintains that speed (lines 5 and 6). If the time remaining to reach the intersection is greater than the time until the traffic light turns green, the policy initiates a gliding maneuver to arrive precisely when the light changes (lines 7, 8, and 9). If neither condition applies, it defaults to natural driving behavior, following the IDM car-following model~\cite{Treiber2000CongestedTS} (lines 10 and 11).

\subsubsection{What makes the nominal policy suboptimal?} 
\label{nominal_policy_limitations}
The nominal policy has inherent limitations due to the simplifications made for real-time feasibility. First, it focuses solely on the ego vehicle's dynamics, ignoring nearby vehicles, which compromises its effectiveness, especially in scenarios with human-driven vehicles, lane changes, or intersection queues.

Furthermore, the policy doesn't account for appropriate vehicle behaviors during unprotected left turns, leading to suboptimal control results and undermining intersection queues and lane changes. Dedicated left turn lanes and traffic signal phases introduce unmodeled lane changes, rendering the policy ineffective when these conditions arise, impacting its emission reduction objective.

\begin{table*}
    \centering
    \begin{tabular}{|c|c|c|c|c|c|c|}
        \hline
        \multirow{2}{*}{Method} & \multicolumn{3}{c|}{20\% penetration} & \multicolumn{3}{c|}{100\% penetration} \\
        \cline{2-7}
        & Emission $\downarrow$ & Speed $\uparrow$ & Throughput $\uparrow$ & Emission $\downarrow$ & Speed $\uparrow$ & Throughput $\uparrow$\\
        \hline
        Multi-task learning & 64.08\% & -27.70\% & -34.70\% & 95.86\% & -30.87\% & -68.11\% \\
        \hline
        Nominal policy & 13.13\% & -21.11\% & -30.07\% & -25.09\% & 11.72\% & -3.90\% \\
        \hline
        Multi-residual task learning (Ours) & \textbf{-13.95\%} & \textbf{12.35\%} & \textbf{7.95\%} & \textbf{-29.09\%} & \textbf{17.10\%} & \textbf{5.72\%} \\
        \hline
    \end{tabular}
    \caption{Performance comparison of MRTL with other baselines for eco-driving at 20\% and 100\% AV penetration. The percentages are calculated compared to the naturalistic human-like driving denoted by the IDM baseline. Evaluation metrics involve emissions reduction and speed improvement of vehicles and throughput improvement at the intersection - where lower emissions, higher speed, and higher throughput percentages indicate better performance.}
    \label{tab:r1-answer}
\end{table*}

\subsection{MRTL Implementation Details}
We employ centralized training and decentralized execution for training MRTL policies. We use actor-critic architecture with three hidden layers, each with 128 neurons in both the actor and critic, with a learning rate of 0.005. 1200 traffic scenarios are modeled in 600 intersections and two AV penetration levels (20\% and 100\%) in the SUMO simulator. PPO algorithm~\cite{schulman2017proximal} is used as the DRL algorithm with 12 workers running for 400 iterations. We use a neural surrogate emission model~\cite{sanchez2022learning} replicating MOVES~\cite{epa_moves} as our emission model to measure vehicular emissions. 

\begin{figure*}[h]
  \begin{subfigure}[b]{0.32\textwidth}
    \includegraphics[width=\linewidth]{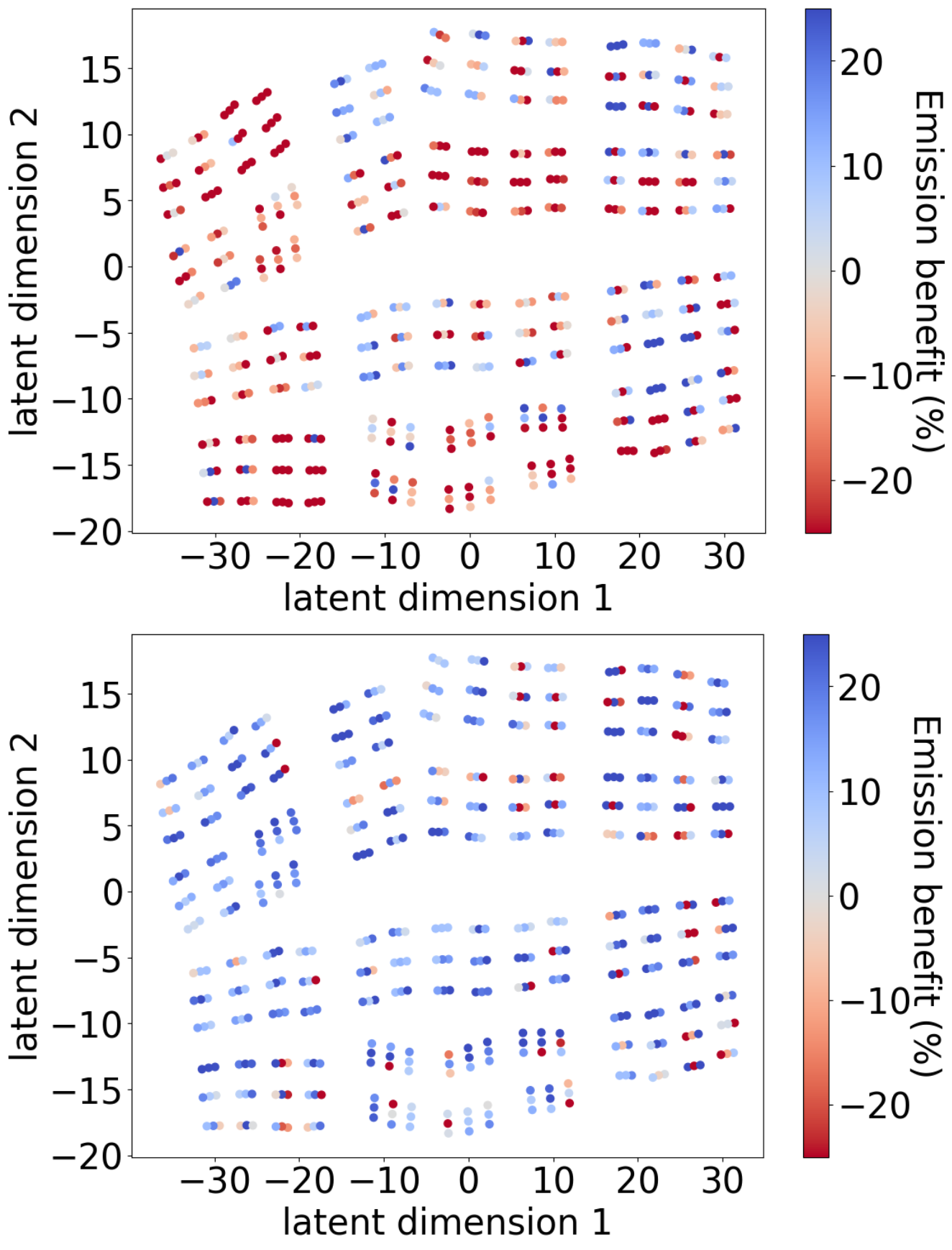}
    \caption{Emision benefits when 20\% penetration of AVs are used with nominal policy (top) and multi-residual task learning (bottom)}
    \label{fig:e1}
  \end{subfigure}
  \hfill 
  \begin{subfigure}[b]{0.32\textwidth}
    \includegraphics[width=\linewidth]{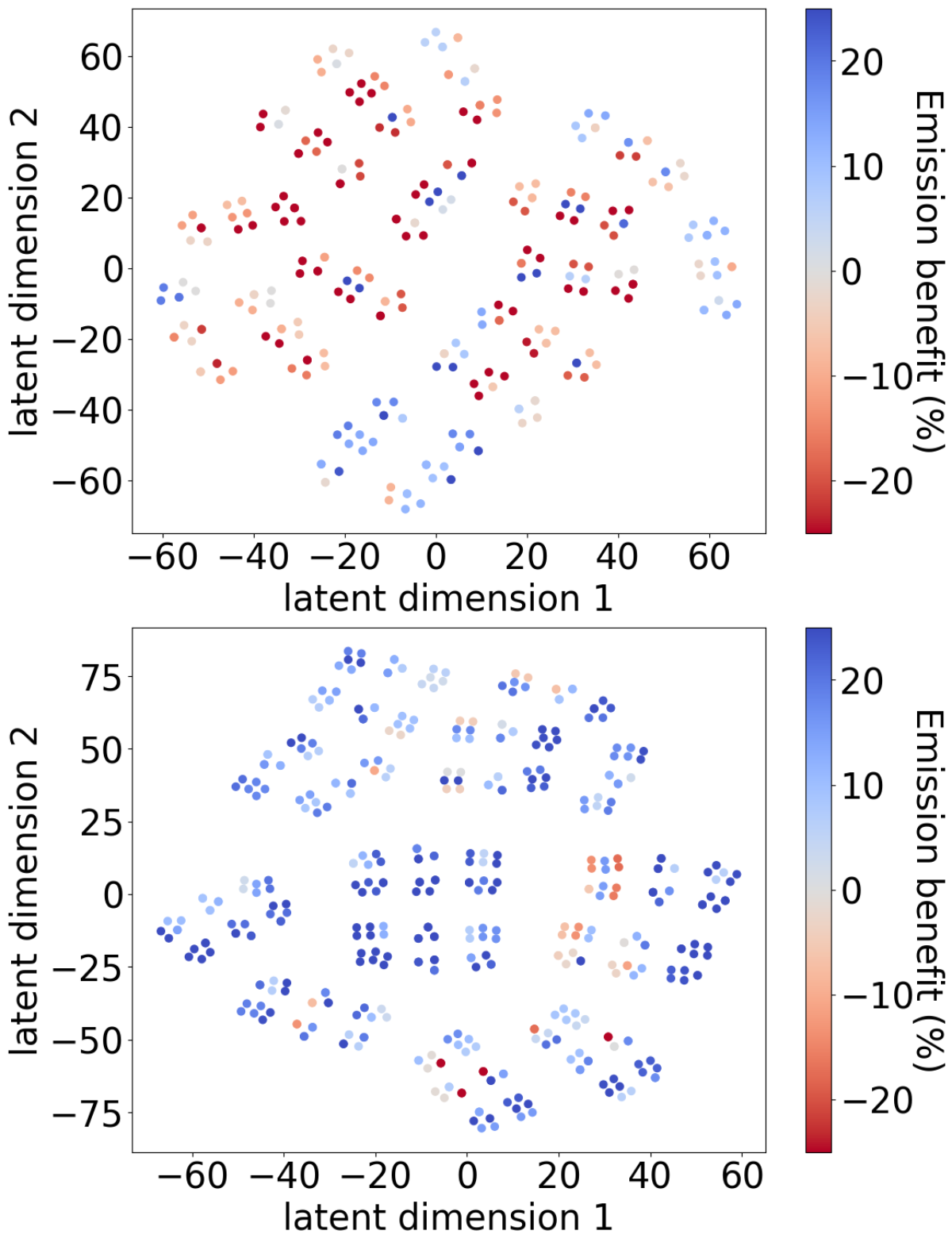}
    \caption{Emision benefits when protected left turns are present with nominal policy (top) and multi-residual task learning (bottom)}
    \label{fig:e3}
  \end{subfigure}
    \hfill 
  \begin{subfigure}[b]{0.32\textwidth}
    \includegraphics[width=\linewidth]{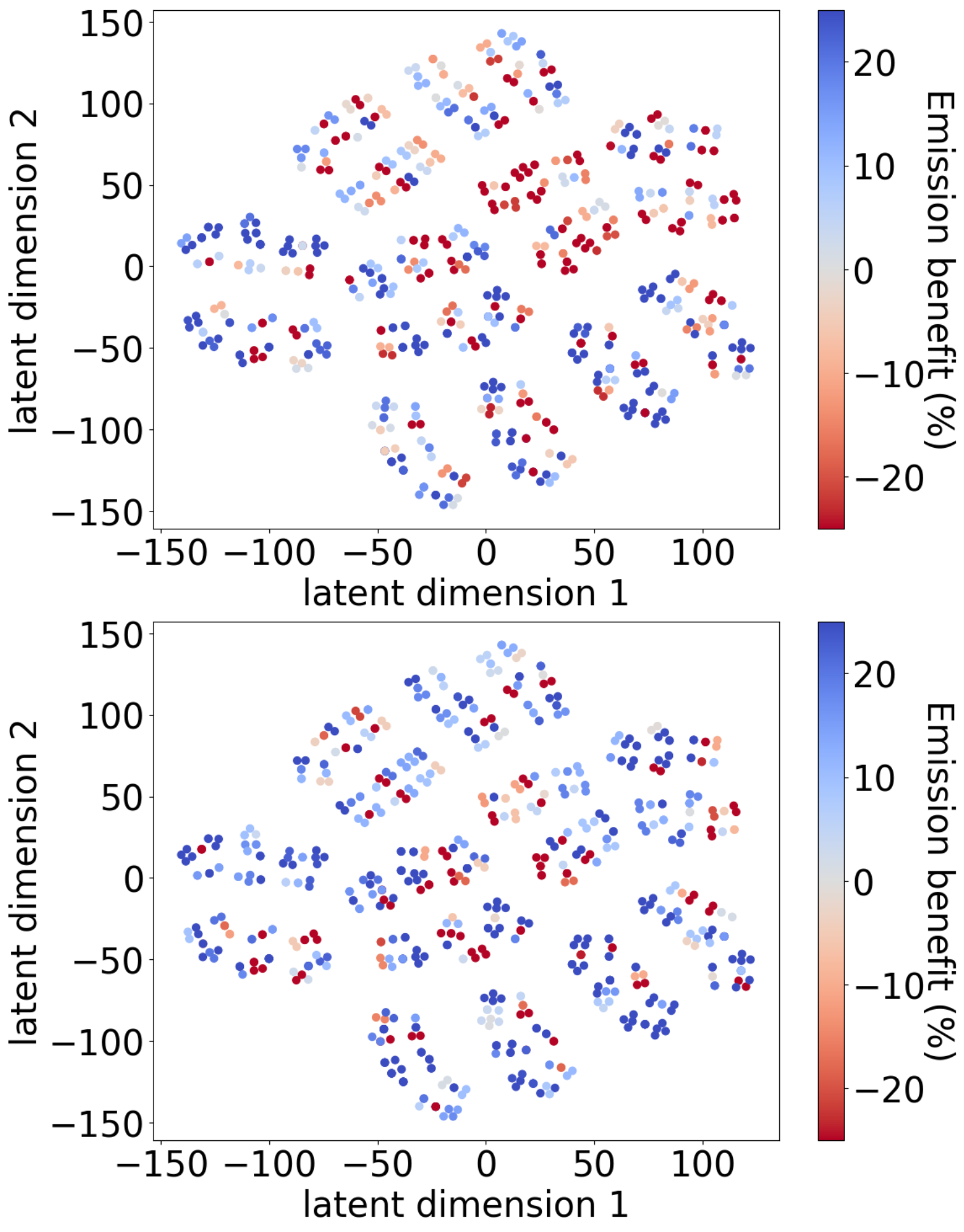}
    \caption{Emision benefits when unprotected left turns are present with nominal policy (top) and multi-residual task learning (bottom)}
    \label{fig:e2}
  \end{subfigure}
  \hfill 
\caption{Visualization of t-SNE plots illustrating emission benefits (higher the better) in assessing the efficacy of MRTL policy in mitigating nominal policy limitations. t-SNE is used for dimensionality reduction of vectors describing incoming approaches to a two dimensional space (latent dimension 1 and 2 in the figures). Thus, each data point is an incoming approach, and the color denotes the emission benefits (a) with partial guided AV penetration (20\%), (b) in the presence of protected left turns, and (c) when dealing with unprotected left turns. In all cases, the MRTL policy outperforms the nominal policy, evidenced by the predominance of blue data points in the lower-row figures as compared to the upper-row figures. \vspace*{-0.5cm}}
\label{fig:r2-results}
\end{figure*}

Each MDP in eco-driving cMDP is defined as follows. 
\begin{itemize}
\item \textbf{States}: speed, position of the ego-vehicle, leading and following vehicles in the same lane and adjacent lanes, the current traffic signal phase with remaining time, and context features including lane length, speed limit, green and red phase times, and approach phase count.
\item \textbf{Actions}: longitudinal accelerations of the ego-vehicles. Lane changes are done by SUMO and not by the policy.
\item \textbf{Rewards}: Ego-vehicle rewards are computed as $v_i(t) + w_1 e_i(t)$, where $w_1=-7.57$ is a hyperparameter, and $v_i(t)$ and $e_i(t)$ represent the ego-vehicle's speed and emissions at time $t$, respectively. We use increasing $v_i(t)$ as a proxy for travel time reduction. 
\end{itemize}

We adopt a neural network initialization method inspired by Silver et al.~\cite{silver2018residual}. Initially, we set the last layer of the policy network to zero (and hence $f_\theta(s, c) = 0$ at the start). This prevents the MRTL policy from being worse than the nominal policy, especially when the nominal policy is close to optimal. We also include a 30-iteration pre-training phase for the critic to align better with the nominal policy, improving value estimates early on.

\section{Experimental Results}

Here, we present the experimental results of employing MRTL for eco-driving at signalized intersections.

\subsection{Baselines}

In order to assess the benefit of the MRTL framework, we leverage three baselines to compare the performance. 

\begin{enumerate}
    \item \textbf{Intelligent Driver Model (IDM)~\cite{Treiber2000CongestedTS}:} human-like driving baseline. The IDM~\cite{Treiber2000CongestedTS} is used.
    \item \textbf{Multi-task reinforcement learning:} Multi-task reinforcement learning from the scratch as introduced in Section~\ref{multi-task-learning}.
    \item \textbf{Nominal policy:} policy in algorithm~\ref{nominal_policy}. 
\end{enumerate}

We do not use exhaustive training (training a different model on each intersection) as a baseline since it is prohibitively expensive given the large number of intersections and, hence, practically less useful for eco-driving.

\subsection{Performance and generalization}

In Table~\ref{tab:r1-answer}, we analyze emission reduction and speed improvements of vehicles and throughput increase at the intersection at 20\% and 100\% AV penetration across 600 signalized intersections. Our findings highlight MRTL's effectiveness in enhancing emission reductions due to better generalization. 
Our MRTL policy improves benefits even in partial penetration scenarios when the nominal policy falls short. Furthermore, training multi-task reinforcement learning policies from scratch is challenging at both penetration levels. Suboptimal individual agent performances in multi-agent settings can lead to training collapse, especially in scenarios where vehicles follow one another. This can be seen in the significant emissions increase in multi-task reinforcement learning when comparing 20\% to 100\% penetration.

\subsection{Nominal policy limitations}

In Section~\ref{nominal_policy_limitations}, we discussed limitations in our nominal policy design. Here, in Figure~\ref{fig:r2-results}, we explore how the MRTL framework effectively addresses these limitations. We focus on three settings: partial penetration, intersections with protected left turns, and those with unprotected left turns. Through t-SNE~\cite{van2008visualizing} distribution plots as performance profiles, we show that MRTL significantly improves over the nominal policy performance in all three settings, with benefits extending across majority of intersections.

\subsection{Control noise and bias noise}

\begin{figure}[h]
\centering
\includegraphics[width=\linewidth]{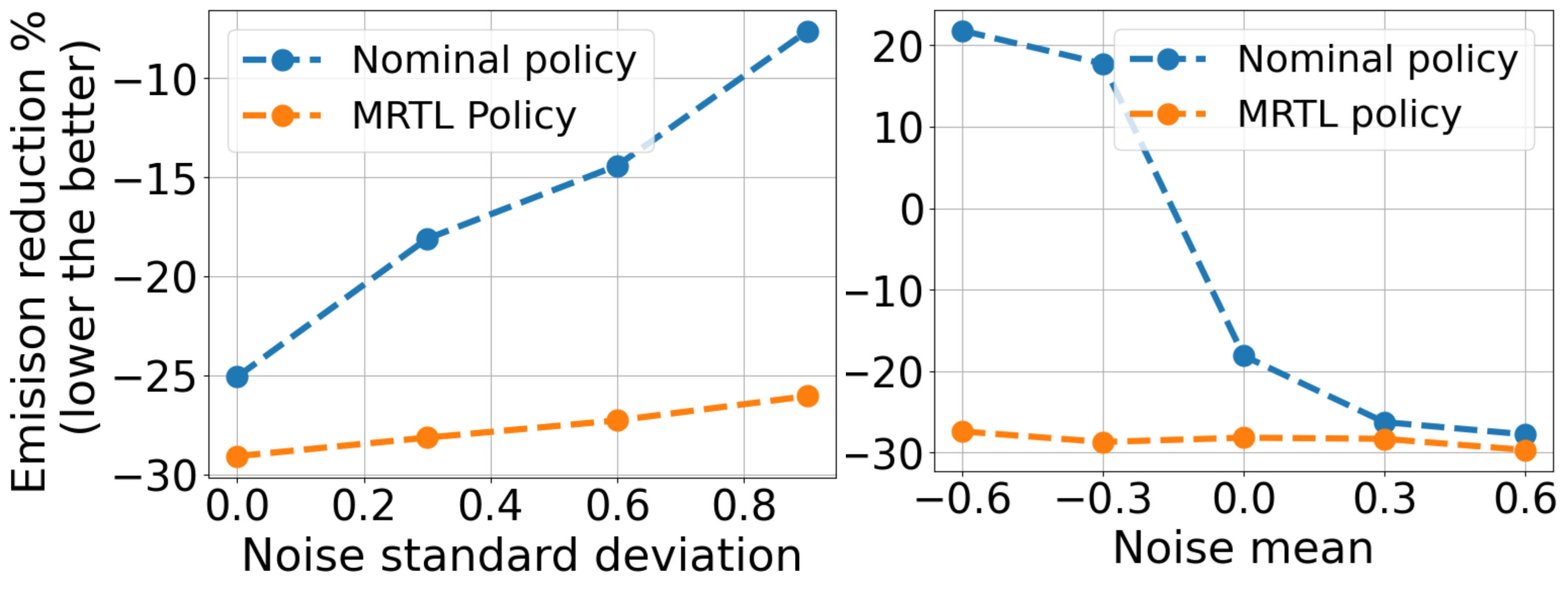}
\caption{Effect of control noise (left) and bias noise (right) on emissions.}
\label{control-noise}
\end{figure}

While conventional eco-driving controllers like GLOSA~\cite{katsaros2011performance} struggle with noise from communication delays and sensor issues, MRTL policies can adapt well to such noise. Moreover, the nominal policies can be biased toward certain cities or conditions, but DRL can learn to adapt on top of them. To test this adaptability, we introduce control gaussian noise $\epsilon_c = \mathcal{N}(0, \sigma^2)$, varying $\sigma^2$ and a bias gaussian noise $\epsilon_b = \mathcal{N}(\mu, 0.3)$, varying $\mu$ to AV accelerations. In Figure~\ref{control-noise} left, MRTL policies are more resilient to control noise, with only a 3\% performance decrease compared to a significant 18\% drop in the nominal policy. Similar results can also be seen under bias noise in Figure~\ref{control-noise} right.

\vspace{-0.5cm}
\subsection{Why does multi-residual task learning work?}

\begin{figure}
    \centering
    \includegraphics[width=\linewidth]{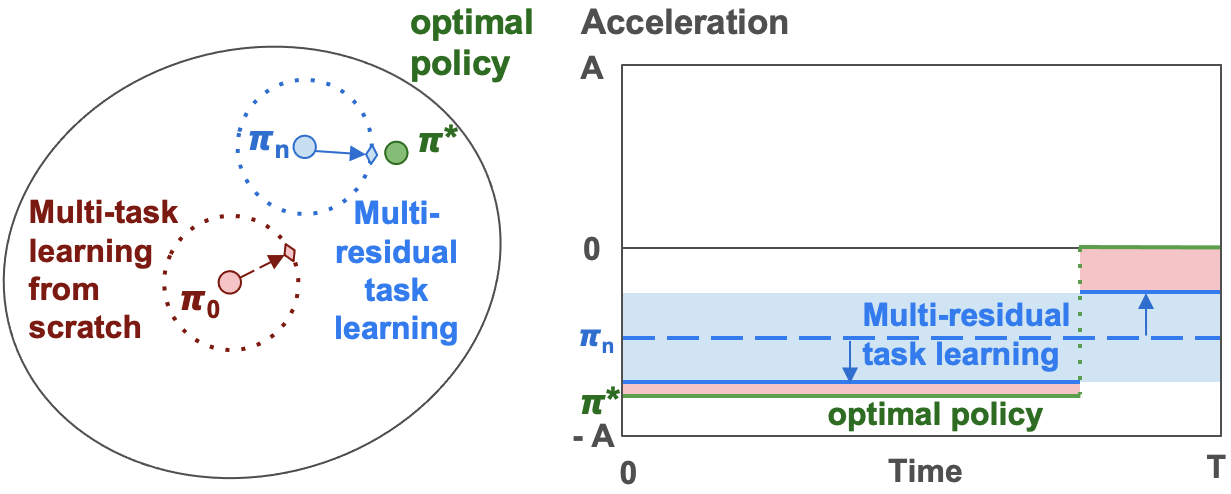}
    \caption{Schematic interpretation of MRTL in policy search. Left: MRTL enables better policy search initialization compared to initializing from scratch. Right: A concrete example from eco-driving at signalized intersections. \vspace*{-0.4cm}}
    \label{fig:laneswitch_notation}
\end{figure}

Intuitively, MRTL simplifies policy search by fine-tuning from a nominal policy that is suboptimal yet not too far from the optimal policy, while learning a multi-task policy from scratch necessitates more computational effort due to possibly distant random initialization. This contrast is illustrated in Figure~\ref{fig:laneswitch_notation} (left) with $\pi^*$ as the optimal policy, $\pi_n$ as the nominal policy for MRTL, and $\pi_0$ as the random initialization. MRTL allows searching within a high-quality ball around $\pi_n$, yielding a good solution near the optimal policy, while $\pi_0$ remains far from optimal under the same computational budget. In practice, the distance between policies and the performance landscape may be highly nonconvex due to the nonconvexity of the objective function and the policy learning process. However, the nominal policy offers a favorable warm start to MRTL by initializing the search closer to the optimal policy compared to random initialization in learning from scratch.

As an example, consider a vehicle that eco-drives when encountering a red light. In Figure~\ref{fig:laneswitch_notation} (right), we contextualize the above interpretation with a commonly observed behavior in a subset of our traffic experiments by considering a general MDP that can lead to multiple MDPs based on traffic signal timing plans. We take the known strategy of gliding (constant deceleration throughout) as the nominal policy $\pi_n$~\cite{huang2018eco}. With potential deviations like human vehicles at the traffic light, the optimal policy $\pi^*$ may involve piecewise-constant acceleration (glide until the leading vehicle is met, then constant velocity to keep a constant headway, generally for a short time period before crossing the intersection).

MRTL from the gliding policy allows the search space (blue region) and the best policy within the search space (blue solid lines) to be close to the optimal policy. In contrast, random initialization from the entire action space $\prod_{t=1}^{T} a_t$, where $a_t \in [-A, A]$, on average results in the zero acceleration policy, which is further away from the optimal policy than the gliding residual initialization. Moreover, random initialization usually leads to non-smooth acceleration profiles in practice, potentially making policy search challenging, whereas the constant deceleration from the gliding policy for MRTL allows a smoother learning landscape.  

\section{Conclusion}

This study examines the algorithmic generalization of DRL in solving contextual Markov decision processes. We present MRTL as a generic framework for achieving this goal. MRTL uses DRL to acquire residual functions, improving upon conventional controllers. We apply MRTL to cooperative eco-driving, showing improved generalization in emission reductions. Potential future work includes analyzing MRTL to further understand the impact of different nominal policies on generalization.  

\addtolength{\textheight}{-8cm}   







\bibliographystyle{unsrt} 
\bibliography{references}


\end{document}